\documentclass[conference,11pt]{IEEEtran}
\usepackage{graphicx}
\usepackage[procnames]{listings}
\usepackage{hyperref}
\def\BibTeX{{\rm B\kern-.05em{\sc i\kern-.025em b}\kern-.08em
    T\kern-.1667em\lower.7ex\hbox{E}\kern-.125emX}}

\begin{document}

\title{PHS: A Toolbox for Parallel Hyperparameter Search}

\author{
\IEEEauthorblockN{1\textsuperscript{st} Peter Michael Habelitz}
\IEEEauthorblockA{\textit{Competence Center for High Performance Computing} \\
\textit{Fraunhofer ITWM}\\
Kaiserslautern, Germany \\
peter.michael.habelitz@itwm.fraunhofer.de}
\and
\IEEEauthorblockN{2\textsuperscript{nd} Janis Keuper}
\IEEEauthorblockA{\textit{Institute for Machine Learning and Analytics (IMLA)} \\
\textit{Offenburg University}, Germany\\
keuper@imla.ai}
}

\maketitle

\begin{abstract}
We introduce an open source python framework named \emph{PHS - Parallel Hyperparameter Search} to enable hyperparameter optimization on numerous compute instances of any arbitrary python function. This is achieved with minimal modifications inside the target function. Possible applications appear in expensive to evaluate numerical computations which strongly depend on hyperparameters such as machine learning. Bayesian optimization is chosen as a sample efficient method to propose the next query set of parameters.\\

\noindent Source code: \url{https://github.com/cc-hpc-itwm/PHS}\\

\end{abstract}


\section{INTRODUCTION}
In recent years, the requirements in the field of machine learning have changed drastically: There is an enlarging amount of data and the computing power is constantly increasing. Research is also dealing with more complex and more difficult questions, such that the associated algorithms are becoming more complicated. This growing complexity also affects the number of hyperparameters to be configured. These range from simple aspects such as regularization parameters to complete design decisions about the model architecture to be used. Obviously, a good choice of the hyperparameters is crucial for the overall performance of the system.\\
Unfortunately, it is very difficult to find an appropriate parameter setting with a trial-and-error tactic. While a good parameter combination can be found by chance, the parameter search is a computationally very intense: The larger the data set and the more complicated the machine learning approach, the longer the evaluation of a single parameter takes. The search for a suitable parameter set must therefore be carried out as efficiently as possible.

\section{CURRENT STATE OF THE ART}
The variety of different approaches to hyperparameter searches is very broad: 
From random search \cite{bergstra2012} to random forests \cite{hutter2011sequential}, to particle swarm optimization \cite{lorenzo2017particle}, up to Bayesian optimization \cite{snoek2015scalable,shahriari2015taking}. Especially the latter is a useful environment for the optimization of expensive functions. It naturally handles the balancing between exploration and exploitation by means of its acquisition function. 

\section{METHODS AND KEY RESULTS}
Many algorithms have some \emph{free} parameters which are set at the beginning of an evaluation. \emph{free} means the impact on the result is in theory and practice not fully understood and can therefor be set in a reasonable range. In machine learning and deep learning these values are called \emph{hyperparameters} whose relevance are explained in the following.
\begin{itemize}
	
	\item Research in that field has proven to be a highly empirically driven task. This means that progress is often rooted in experiments which showed improvements in some specific aspect. Thereafter researchers hop onto and try to find theoretical explanations. One nice example is (batch normalization). This is not a bad situation in general but it emphasizes the lack of a broad and profound mathematical foundation like in other disciplines for instance in physics. The missing theory is an explanation for the question why the hyperparameters and no rules for their choice exist.
	\item Practitioners worry about high performance when implementing machine learning algorithms. Every step in the workflow from data preparation and augmentation, over the classical hyperparameters like batch size, learning rate, ... up to the neural network design can be tuned. Hyperparameter optimization is crucial and requires a well organized handling to not get lost.
	\item Hyperparameter optimization can be trivially parallelized. This means that the algorithm can be handled as a black box which is evaluated with a certain set of parameters. Some internally defined measure like validation loss or prediction accuracy is the only feedback and used to assess the used parameter set. The evaluations (trainings) of multiple black boxes are independent from each other once the parameter set is fixed. As long as a single compute instance (e.g. GPU or node) has enough memory and does not need to much time carrying out the computation, there is no demand for any complex parallelization scheme which distribute one black box evaluation over multiple instances. In addition there is no more efficient way of parallelization because very little communication is necessary.
\end{itemize}

\subsection{Description of the framework}
As mentioned above selecting a particular set of hyperparameter is crucial for most machine learning algorithms. That said it implies a highly repetitive task for both, researchers and engineers. To take off the implementation work of hyperparameter searches we developed a framework named \emph{phs - parallel hyperparameter search}. In order to reach this goal we concentrated on some key features.

\begin{itemize}
    \item \textbf{parallelization} The framework takes advantage of the simple communication scheme needed to parallelize each function evaluation. After a simple definition of your workers you can fully utilize a cluster of compute nodes.
    \item \textbf{ergonomics} As the framework introduces no algorithmic novelties the gain comes with an easy to use abstraction of the hyperparameter search. Therefor it is important to provide a clear and precise API.
    \item \textbf{generality} There is no binding to a particular python framework like TensorFlow or PyTorch. A blackbox approach enables to perform hyperparameter search on every python function. Random search is capable of all kinds of parameter types (continuous and discrete numerical, categorical, arbitrary python statements) while Bayesian optimization needs continuous numerical parameters.
    \item \textbf{documentation} A full documentation from installation and quick start guide over examples down to detailed explanations of the modules is available [1].
\end{itemize}

\subsection{Implemented Search Strategies}
There is a wide variety of search strategies to deal with hyperparameter optimization in the domain of machine learning. \emph{phs} supports manual (explicit) search, random search and Bayesian optimization with Gaussian processes.

Manual search is probably a widespread habit to some extend and can lead to acceptable results based on the individual experience. Grid search and random search are conceptional simple to automate, while random search usually is more efficient. The reason for that is a high variability in the importance of different parameters. As a consequence grid search spend too many trials in irrelevant regions \cite{bergstra2012}. Because of its simplicity and no need for any additional computation, random search is well suited as a base line to compare with other strategies and as an initialization for Bayesian optimization \cite{mockus1974}.

Having in mind the exponential growth of the search space with the number of hyperparameters and the potentially high cost for one evaluation the demand for an efficient choice which parameter values are worth for probing are obvious. Here Bayesian optimization comes into play. It treats the black box as a random function and tries to describe it with a probabilistic surrogate model, mapping from parameter space to the black box result. The characteristic of each Bayesian method is to use a prior distribution over functions. It is updated after each query of the black box forming the posterior distribution over functions (the surrogate model). One possible assumption for a prior is a Gaussian process \cite{rasmussen2006}, which is defined with its mean and covariance function (also called kernel). The last missing part is the choice of an acquisition function, which determines the next query point based on the posterior distribution of functions. Due to the probabilistic nature it is always a trade off between exploration (Gaussian process has a high uncertainty about the result of the black box at the next query point) and exploitation (Gaussian process has a low uncertainty about the result and is close to a previous minima of the black box at the next query point).

\subsection{Parallelization Technique}
At the moment the usage of two different underlying parallelization libraries are implemented, Dask[2] and python local processes. These share the same functionalities but differ in definition of workers and technology of task scheduling. On the software side there is one lightweight master running on the machine from which the experiment is started. It executes the parameter setup, manages the task scheduling to the workers and gathers some of the results immediately as they are completed.

Each client acts as a worker and is charged with the evaluation of the target function of interest. Even the Bayesian optimization for suggesting new parameter values is done on the workers themselves. By this means the master is relieved and the prerequisites for a solid scaling behavior are ensured.

\subsubsection{DASK}
Dask \cite{dask} is a flexible library for parallel computing in Python. It provides dynamic task scheduling and 'Big Data' collections. Hereof futures are enough to schedule tasks to the workers. Dask is meant to be the parallelization back end for productive usage.

\subsubsection{Processes}
Beside Dask native processes of the Python built-in concurrent.futures module is implemented as an alternative back end. It provides the same functionalities and user experience. Local processes serve as workers, which means that the CPU cores of one machine can be utilized. The intention of this kind of computing resources is less on the computation heavy use in terms of production but rather on testing and debugging especially when no HPC system is available. But taking CPU only function evaluations into account, the processes version can also be utilized in a meaningful manner.

\subsection{Quick Start}
\begin{figure}[h!]
  \centering
    \includegraphics[width=\linewidth]{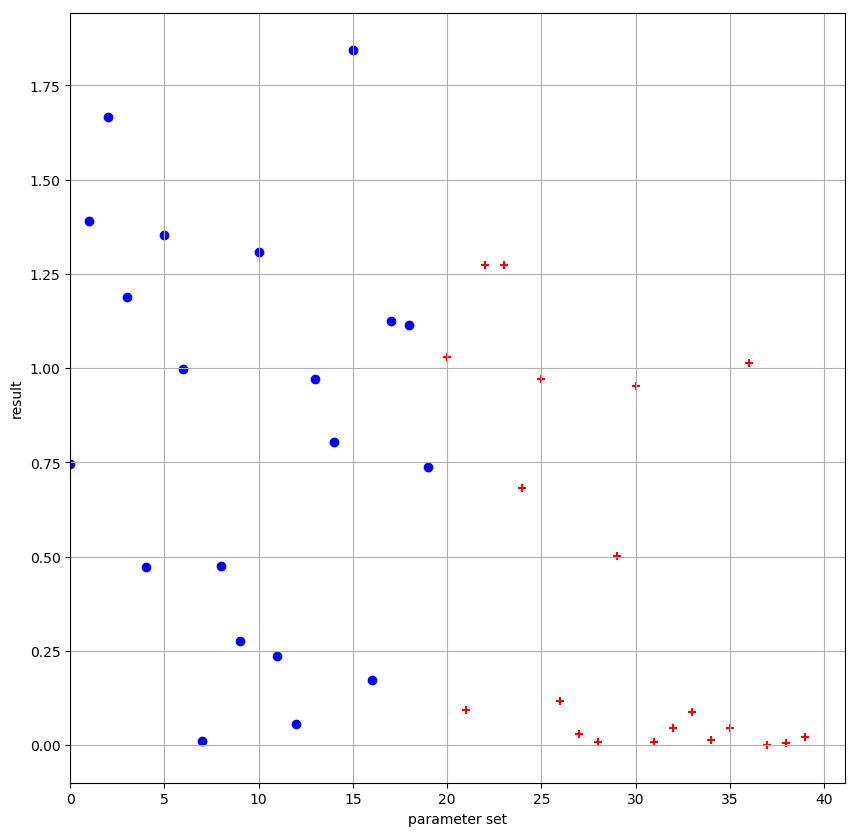}
    \caption{The returned value of the Griewank function is plotted over the parameter set index. Blue dots represent sets generated by random choice, red crosses indicate involved Bayesian Optimization. It is to be seen how the acquisition function suggest parameter values based on exploration and exploitation.}
    \label{fig:resultsQuickStart1}
\end{figure}

The easiest way to become familiar with the tool is to go through the following example of finding minima of the second order Griewank function, which is a common test scenario for optimization algorithms. All three generated visualizations are shown in figure \ref{fig:resultsQuickStart1}.

\begin{enumerate}
    \item \textbf{Preparation of the target function}
    \item \textbf{Definition of the search strategy}
    \item \textbf{Experiment setup}
    \item \textbf{Computation setup}
    \item \textbf{Post process the results}
\end{enumerate}

\begin{figure}[h!]
  \centering
    \includegraphics[width=\linewidth]{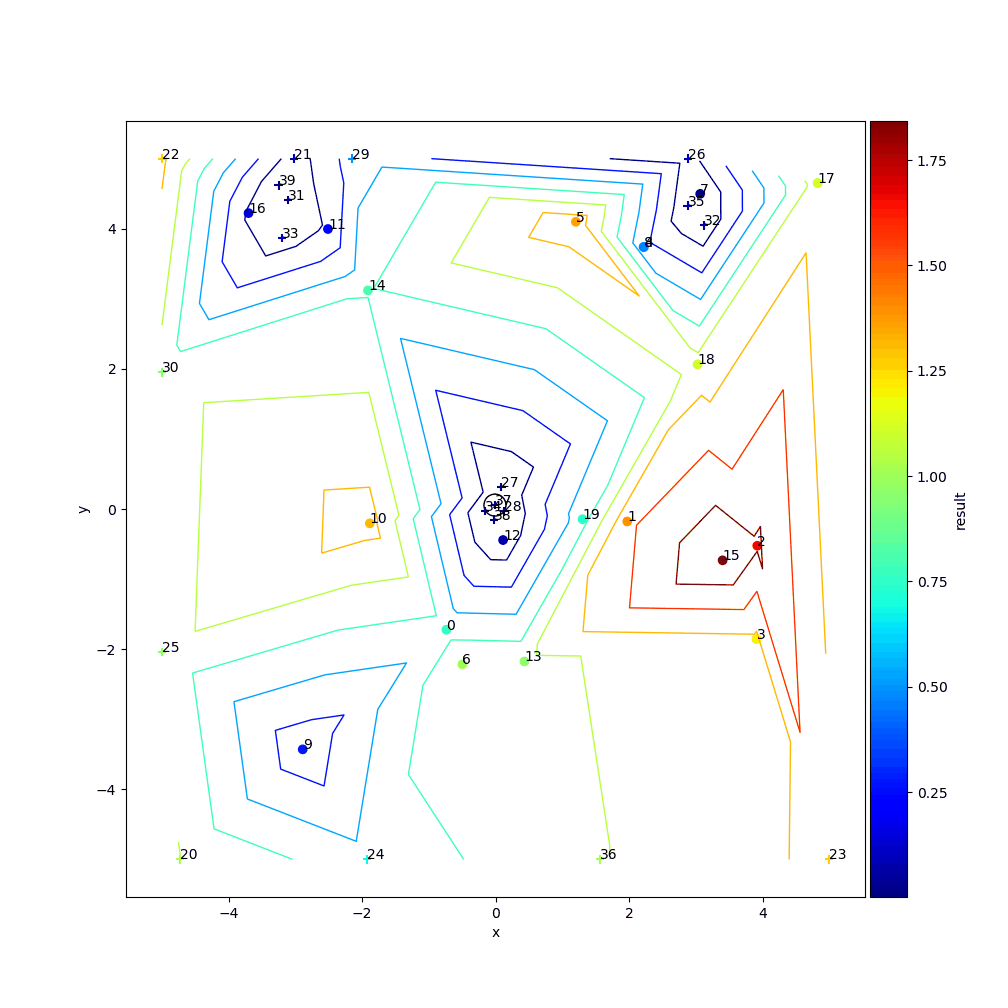}
    \caption{Parameter x and y are plotted with the color decoded result. The numbers are the indices of the set. Crosses indicate involved Bayesian Optimization. The blue circle shows the best (lowest) result.}
    \label{fig:resultsQuickStart2}
\end{figure}

\begin{figure}[h!]
    \centering
    \includegraphics[width=\linewidth]{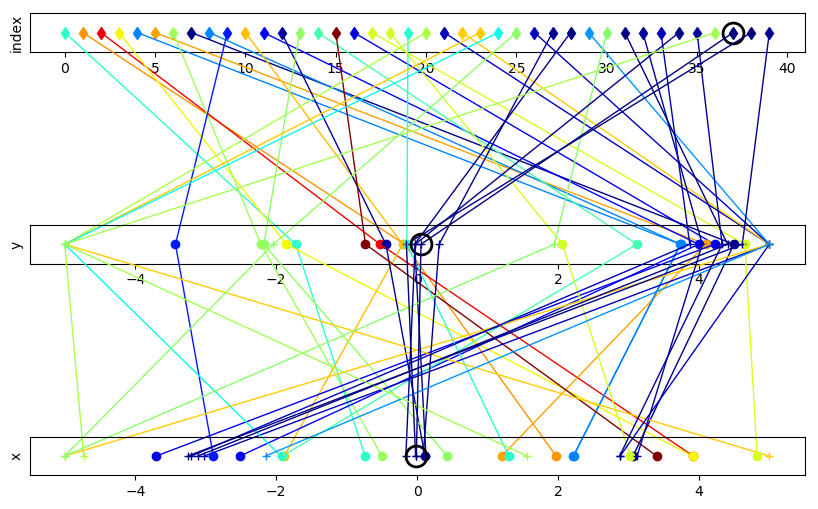}
    \caption{For each parameter defined in the experiment, here x and y, a separate one dimensional scatter plot is drawn. The points belonging to the same parameter set are connected via lines. These lines show the color coded result. This diagram can be useful to see combinations of parameter values which yield good results in case of high dimensional search spaces.}
    \label{fig:resultsQuickStart3}
\end{figure}

\newpage

\section{cifar10 Example}
Here we show an example for the application of phs on image classification with deep learning using the cifar10 data set. The intention is to provide a showcase to help users applying a hyperparameter search on their own DL tasks. This experiment is not meant to optimize for a good hyperparameter set what would require more careful work on the details.

In this experiment learning rate, batch size and weight decay are considered as hyperparameters. As initialization 10 parameter sets are selected randomly. Thereafter 20 sets with Bayesian optimization follow. Here learning rate and weight decay are optimized. The result of such an experiment should not only consist in the extraction of the best parameter set as there is much more possible gain of insights. To help you examine the results, the post-processing module from phs can be applied to the experiment.

\begin{figure}[h!]
    \centering
	\includegraphics[width=\linewidth]{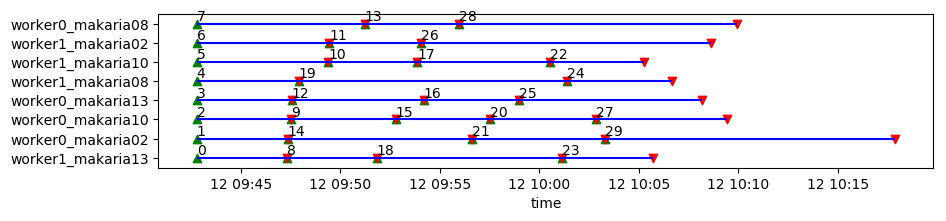}
    \caption{Worker time line for each of the 30 function evaluations (cifar10 trainings) on the involved compute nodes. Green triangles mark the starts and red triangles the ends of each computation.}
    \label{fig:cifar10wt}
\end{figure}

\begin{figure}[h!]
	\centering
	\includegraphics[width=\linewidth]{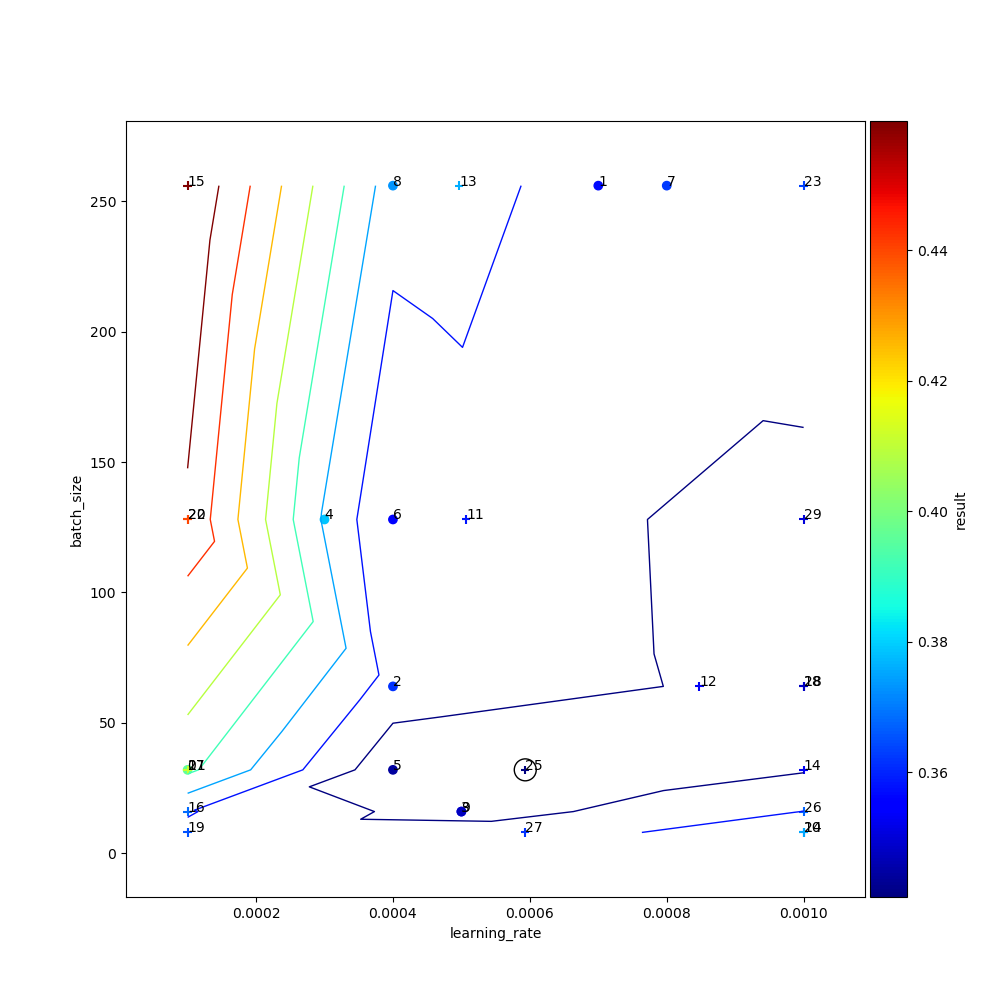}
	\caption{Interpolated contour plot of batch size over learning rate with the color coded result. The blue circle shows the best (lowest) result.}
	\label{fig:cifar10lrbs}
\end{figure}

\begin{figure}[h!]
	\centering
	\includegraphics[width=\linewidth]{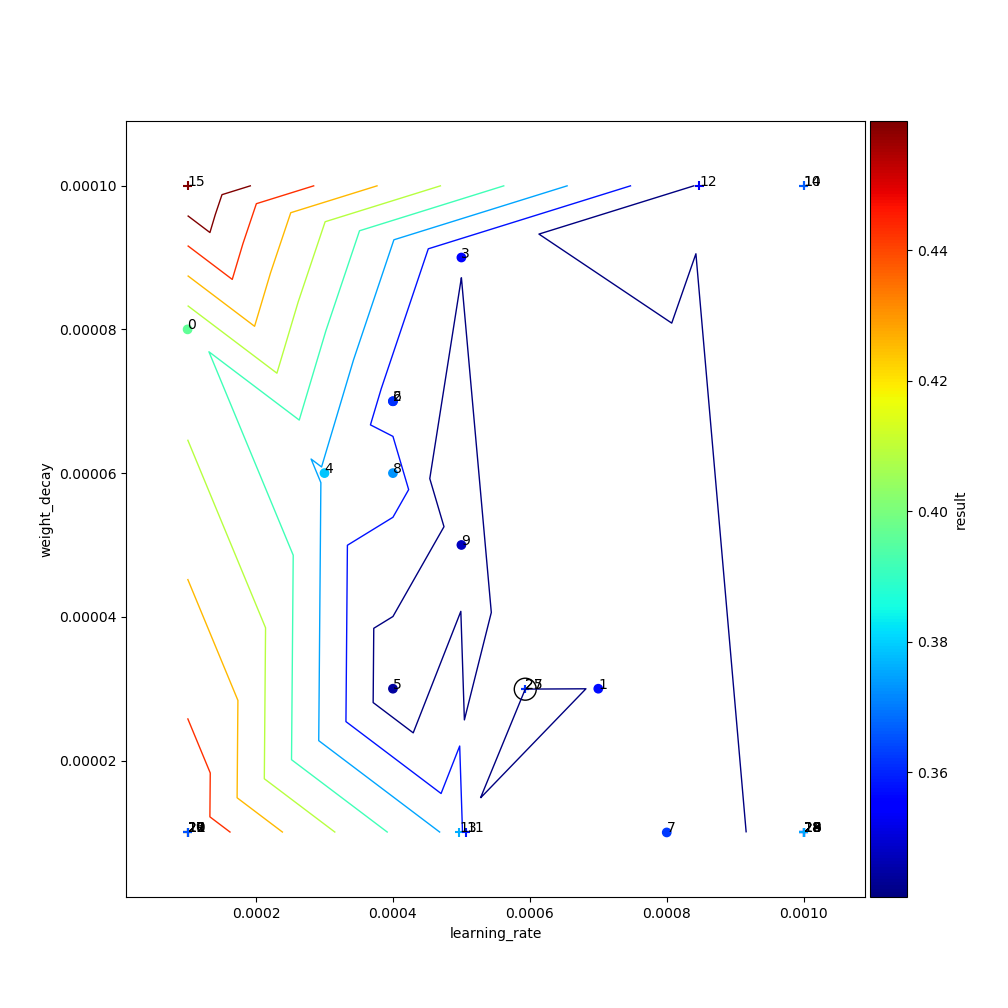}
	\caption{Interpolated contour plot of batch size over weight decay together with the color coded result. The blue circle shows the best (lowest) result.}
	\label{fig:cifar10lrwd}
\end{figure}

In figure \ref{fig:cifar10wt} you see the time consumption of each query. Contour plots of batch size and weight decay over learning rate are shown in figure \ref{fig:cifar10lrbs} and \ref{fig:cifar10lrwd}.

\clearpage

\section{CONCLUSION}
Hyperparameter search in the domain of Deep Learning remains challenging even with supporting tools.

\begin{itemize}
    \item Deep Learning algorithms easily give rise to look into 10s of hyperparameters which span a huge search space. This would require to evaluate many different sets which contradicts to the high computational cost of a single run.
    \item In the real world hyperparameter search is not an automatism at all. There are many design choices upon the hyperparameter search itself. It starts with the choice which variables should be declared as hyperparameters and how to set the range for them and continues to the search strategy itself.
    \item Another very important field is the evaluation of the results. Simple statements in a universal meaning is what we are looking for like 'parameter x of value 0.01 works best'. But these are most of the time very difficult to make because the experiments are so complex. Even when declaring many hyperparameters there will be many variables fixed. To asses a result, every detail has to be taken into account and cannot be isolated. Every detail has to mentioned.
    \item Optimization of a neural net includes some randomized processes like the initialization of the weights. This leads to different results when evaluating the same parameter set multiple times. As a consequence it is recommended to always run for example 3 experiments and report the average as the result of this parameter set.
\end{itemize}

\newpage

\section*{Acknowledgements}
This work has been funded by the Federal Ministry of Education and Research (BMBF)	Germany, as part of the "Deep Topology Learning" (DeToL) project (see www.detol.de).


\bibliographystyle{IEEEtran}
\bibliography{references}
\end{document}